# Leveraging ChatGPT for Sponsored Ad Detection and Keyword Extraction in YouTube Videos

Brice Valentin Kok-Shun
*Department of Information Systems and Operations Management*
University of Auckland
Auckland, New Zealand
0000-0001-9923-5042

Johnny Chan
*Department of Information Systems and Operations Management*
University of Auckland
Auckland, New Zealand
0000-0002-3535-4533

*Abstract*—This work-in-progress paper presents a novel approach to detecting sponsored advertisement segments in YouTube videos and comparing the advertisement with the main content. Our methodology involves the collection of 421 auto-generated and manual transcripts which are then fed into a prompt-engineered GPT-4o for ad detection, a KeyBERT for keyword extraction, and another iteration of ChatGPT for category identification. The results revealed a significant prevalence of product-related ads across various educational topics, with ad categories refined using GPT-4o into succinct 9 content and 4 advertisement categories. This approach provides a scalable and efficient alternative to traditional ad detection methods while offering new insights into the types and relevance of ads embedded within educational content. This study highlights the potential of LLMs in transforming ad detection processes and improving our understanding of advertisement strategies in digital media.

## I. Introduction

In recent years, video-sharing platforms like YouTube have become dominant sources of entertainment, education, and information [1]. YouTube is invaluable for content creators, marketers, and advertisers. One of the key features of YouTube's revenue model is the integration of sponsored advertisement (ad) segments, which allows content creators to monetize their videos while providing advertisers a direct route to target specific audiences [2]. This model differs from traditional ads that do not involve the content creator and tend to interrupt the video. Sponsored ad segments involves the promotion of the product by the creators themselves during their video content.

From an advertiser's perspective, delivering ads that align with viewers' interests significantly enhances the likelihood of conversion, brand recall, and user engagement. However, the sheer scale and diversity of content on platforms like YouTube present challenges in analyzing and ensuring content-relevant advertisements. Moreover, while much of the focus in the digital marketing landscape has been on targeted advertising algorithms that use demographic data or user behavior [3], there has been limited exploration of the impact of sponsored ad segments.

The main challenge in studying the relationship between advertisements and video content is accurately detecting ad sections. Traditional methods of detecting video ads often rely on audio and video streams [4], [5], [6]. These methods, while powerful, are often computationally intensive. Additionally, the high variability in ad formats from explicit commercials to subtle sponsorship mentions further complicates detection using conventional techniques.

To address these challenges, this paper proposes a novel methodology that leverages Large Language Models (LLMs) to detect and classify ad sections in video transcripts. LLMs, such as OpenAI's GPT models, have demonstrated remarkable capabilities in understanding context, semantics, and nuanced textual patterns, making them ideal candidates for detecting advertisements based on linguistic cues. By identifying and extracting ad sections within video transcripts, we can perform a comparative analysis of the keywords in ads and non-ad sections, allowing us to explore how advertisements align with the overall video content.

We created a dataset of approximately 421 YouTube video transcripts, covering a wide range of categories and topics. Both auto-generated and manual transcripts were collected to ensure diversity and richness in the data. GPT-4o was prompted to identify advertisement segments, keywords were extracted using the KeyBERT model before being refined into succinct categories using GPT-4o.

This study is significant for several reasons. First, it provides a scalable and automated solution for detecting and analyzing advertisements within video content, which is typically labor-intensive when done manually. Second, it offers insights into the relationship between advertisements and video content, which can have profound implications for advertisers seeking to improve targeting strategies and for content creators aiming to optimize sponsored ad placements within their videos. Third, the research lays the groundwork for future advancements in content-based advertising, where the alignment between ad messaging and content themes can be refined using advanced natural language processing (NLP).

In the following sections, we provide a background on ad detection and content analysis in video platforms. We then describe the methodology used to collect and process video transcripts, detect ads, and extract keywords. We subsequently present the results of our analysis. Finally, we discuss the implications of these findings for the future of content-based advertising, the limitations, and continued work.

## II. Background

In this section, we provide an overview of the video content landscape and its challenges, followed by a discussion on the role of LLMs in text analysis, and the current state of keyword extraction and ad detection techniques relevant to YouTube ads.

### A. Video Content Landscape and Challenges

Video platforms like YouTube host an immense and ever-growing content repository covering a broad spectrum of categories, including entertainment, education, gaming, news, and more [1]. The sheer volume and diversity of content present significant challenges for automated analysis. As advertisers seek to reach specific audiences, ensuring that advertisements are contextually relevant to the video content they accompany becomes increasingly important [7], [8]. Content creators also rely heavily on ads as a revenue source, making the successful integration of ads into videos crucial for maintaining viewer engagement and generating income.

The complexity of modern advertising formats poses unique challenges. Advertisements are no longer restricted to explicit commercials before, during, or after video segments. Instead, content creators often include more subtle forms of advertising, such as product placements, sponsorship mentions, or native ads embedded within the video's dialogue [9]. These more nuanced forms of advertising can be difficult to detect using traditional methods, which often rely on metadata, timestamps, or predefined tags.

This challenge is compounded by balancing ad relevance with viewer engagement. Irrelevant or intrusive ads can lead to viewer dissatisfaction and decreased engagement, while highly relevant and well-integrated ads can enhance the viewing experience [10], [11]. Therefore, understanding the relationship between video content and ad content is critical for optimizing ad placement and improving the overall effectiveness of digital marketing strategies.

### B. Large Language Models

The advent of LLMs has significantly transformed the field of NLP, allowing for more sophisticated and context-aware text analysis. LLMs such as OpenAI's GPT-4 have been trained on massive amounts of data, enabling them to capture deep semantic relationships, contextual meanings, and subtle cross-domain linguistic patterns. These models excel in tasks such as text generation, summarization, sentiment analysis, and translation, providing a powerful tool for analyzing and understanding human language at scale.

GPT-4 is built on the transformer architecture [12], which utilizes attention mechanisms to process and generate text. This architecture enables GPT-4 to generate coherent and contextually appropriate responses by weighing the significance of different words and phrases based on their relationships within the input text. One of the key innovations in GPT models is their pre-training and fine-tuning approach. During pre-training, the model learns from a diverse dataset of text drawn from the internet, allowing it to build a robust understanding of language patterns.

Another crucial aspect of working with GPT-4 is prompting, which involves providing the model with specific instructions or contexts to guide its output. Effective prompting is essential to obtaining high-quality results, as the model's responses can vary significantly based on how a prompt is formulated [13]. The process includes several strategies, such as using explicit questions, setting clear contexts, and providing examples of desired outputs [14]. This technique of prompt engineering helps the model understand the nuances of the task at hand and generates outputs more aligned with the researcher's objectives.

The integration of LLMs has also facilitated increased development speed across various applications [15]. The ability to rapidly generate and test hypotheses, refine approaches, and iterate on analysis strategies allows developers and researchers to streamline their workflows. This efficiency is particularly valuable when working with complex datasets or when the project's goals require quick adaptations to methodologies based on emerging insights.

### C. Keyword Extraction and Ad Detection

Keyword extraction is a foundational technique in text analysis, designed to identify the most important or representative terms from a body of text. Traditional approaches such as term frequency-inverse document frequency (TF-IDF) focus on the frequency of terms within a document relative to a larger corpus, helping to highlight statistically significant words. However, these methods often overlook the contextual meaning of words, making them less effective when dealing with nuanced or domain-specific content [16]. More advanced approaches, such as BERT-based models, have emerged to address these limitations by capturing the meaning and context of words within a document [17].

One such model is KeyBERT [18], a keyword extraction model built on BERT embeddings. KeyBERT is particularly useful because it considers the frequency of words and the surrounding context, providing more accurate and relevant keyword extractions. This is critical when dealing with video content, where the dialogue may shift between different topics, and ads may be subtly embedded within the discussion.

Ad detection has traditionally relied on audio and video streams [4], [5], [6], which can be computationally expensive and labor-intensive, especially for large datasets. With the increasing use of embedded advertising formats, detecting ads using traditional methods has become even more challenging. Many current approaches focus on surface-level indicators, such as the presence of commercial breaks or sponsored content labels. However, these methods often fail to detect subtler forms of advertising, where the promotion of a product or service is integrated directly into the dialogue or narrative.

### D. Related Works

YouTube ad detection has traditionally relied on multimedia signal processing, which analyzes audio and video to identify ads [4], [5], [6], [19], [20], [21]. While effective, these methods require multiple complex models to be developed, making them computationally expensive.

NLP-based approaches have struggled with conversational speech, unclear word boundaries, overlapping dialogue, and colloquial language, making ad detection in transcripts difficult [21]. However, LLMs like ChatGPT offer a solution by understanding context and handling conversational nuances, enabling more precise ad identification without the heavy computational load.

Deep learning methods such as LSTM [22] and BERT [23] have been used to identify sections in YouTube videos. These include topic segmentation of videos using visual and textual data. These methods use text descriptions of the video to separate the content into sections based on the subject. These methods rely on generating summaries using fine-tuned models to detect segments. Such methods are computationally expensive and require extensive development cycles. There exists the opportunity to use Gen AI models like ChatGPT which provide a simpler

Sponsored ad segments on YouTube are under-researched, with most studies focusing on and advocating for transparency [24]. Few have explored the types and relevance of ads within video content, an area this study seeks to address using LLMs.

### III. NOVELTY

The novelty of this research lies in its innovative application of LLM for detecting and classifying sponsored ad segments directly from video transcripts, moving beyond traditional methods that rely on metadata or manual annotations. This approach enables the identification of both explicit and subtle ad formats based on linguistic cues, offering a more context-aware solution.

Additionally, the research introduces a unique focus on keyword extraction using `KeyBERT` to compare the thematic alignment between ad sections and non-ad sections of the video transcripts. This allows for a deeper analysis of how well sponsored ad segments align with the content they accompany, a relatively unexplored area in previous studies.

Creating a dataset of 421 video transcripts spanning multiple categories, along with labeled ad and non-ad sections, adds further novelty by providing a more generalizable basis for analysis. Moreover, the study's emphasis on exploring the relationship between ad content and video topics offers new insights into the effectiveness of content-based advertising strategies, making it a significant contribution to the field.

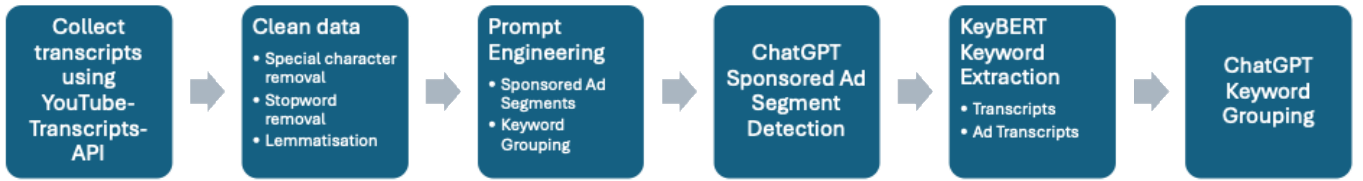

Fig. 1. Ad segment detection and keyword extraction methodology.

## IV. METHODOLOGY

This research employs a systematic approach to create a dataset of video transcripts annotated with sponsored ad segments and relevant keywords, as depicted in Figure 1.

### A. Data Collection

Auto-generated and manual video transcripts are collected from different educational YouTube channels using the `youtube-transcripts-api` Python package.

Selecting multiple channels ensures a comprehensive dataset that reflects different educational topics, teaching styles, and formats. This diverse selection enhances the robustness of our analysis, allowing for a more thorough examination of ad content and keyword extraction across varied contexts within the educational landscape on YouTube.

### B. Text Preprocessing

Prior to ad detection and keyword extraction, we implemented standard preprocessing steps to enhance the quality of the transcripts. This includes (1) Special character removal to ensure that the analysis focuses solely on the words, (2) Stopword removal to reduce noise and focuses the analysis on more meaningful terms, and (3) Lemmatization to ensure that variations of a word are treated as a single entity, improving the consistency of keyword extraction and analysis.

### C. Prompt Design

A carefully crafted prompt was designed to guide the LLM in identifying sponsored ad segments within the transcripts. This prompt was tailored to elicit precise responses regarding the presence of ads, ensuring that the GPT accurately understands the context and intent.

Similarly, a prompt was designed to process the keywords extracted during the Keyword Extraction step. This prompt was developed to aggregate the different keywords into broader categories retaining the original semantic meanings.

### D. Sponsored Ad Segment Detection

The collected transcripts were sent to OpenAI's `gpt-4o-2024-08-06` model to identify ads. The model's performance was evaluated in terms of its ability to detect sponsored ad segments accurately, providing insights into the effectiveness of different LLM architectures in this task.

This evaluation was done by manually checking a small sample of videos and identifying the sponsored ad segment.

### E. Keyword Extraction

Keywords are extracted from the transcripts using the `KeyBERT` Python package. This method leverages BERT embeddings to identify contextually relevant keywords, ensuring that the extracted terms are related to the content of the transcripts. The keywords identified by KeyBERT were then passed through `gpt-4o-2024-08-06` to group them into higher-order categories to reduce dimensionality.

### F. Comparative Analysis

The ad and non-ad sections of video transcripts were compared to assess how sponsored ad segments impact keyword relevance and extraction. For videos with ads, we examine whether ad-related keywords align with the video's theme or introduce unrelated topics. This comparison provides insights into the relationship between ad and video themes.

## V. RESULTS

This section presents the findings from our ad detection and keyword extraction processes, highlighting key insights derived from the analysis of the dataset.

### A. YouTube Sponsored Ad Segments Dataset

The dataset contains 421 transcripts (243 auto-generated transcripts and 178 manually created) collected from 623 videos posted by 6 channels. This is summarized in Table 1.

TABLE I. DATASET SUMMARY

| Channel | Transcripts Collected | |
|---|---|---|
| | *Generated* | *Manual* |
| 3Blue1Brown | 9 | 49 |
| DamiLee | 48 | 9 |
| Fireship | 47 | - |
| Johnny Harris | 48 | 44 |
| PBS Space Time | 44 | 48 |
| SciShow | 47 | 28 |
| **Total** | **243** | **178** |

## B. Prompt Design

Figure 2 shows the prompt was used to identify the ads and Figure 3 shows the one used to group the keywords identified by the KeyBERT algorithm into succinct categories. The iterative prompt design strategy implemented yielded the best results when the format was explicitly stated and instructions were clearly laid out regarding the task and repeated after.

```
Find the ad.
Return in format the following format:
[
  { 'text': str,
    'start': float,
    'duration': float
  },
  ...
]
IF THE AD IS SPLIT ACROSS MULTIPLE DICTIONARIES,
RETURN A LIST OF DICTIONARIES. ONLY RETURN THE
DICTIONARIES IF THE AD IS PRESENT.
If there is no ad, return None.
```

Fig. 2. Prompt for ad segment detection.

```
I will provide you with a list of keywords.
Group the common keywords - create new keywords that describe
each group.
Return in the following format:
[groupkeyword1, groupkeyword2, ...]
FOLLOW THE FORMAT. DO NOT DEVIATE FROM THE FORMAT.
```

Fig. 3. Prompt for keyword grouping.

## C. Sponsored Ad Segment Detection

The ad segment detection identified 109 and 101 ads in the auto-generated (45% of the videos) and manually-created transcripts (57% of the videos). The prevalence of the ads in each channel's videos is shown in Table 2.

TABLE II. SPONSORED AD SEGMENTS PREVALENCE

| Channel | Sponsored Ad Segments Detected | | Prevalence | |
|---|---|---|---|---|
| | *Generated* | *Manual* | *Generated* | *Manual* |
| 3Blue1Brown | - | 3 | 0% | 6% |
| DamiLee | 14 | 7 | 29% | 78% |
| Fireship | 10 | - | 21% | - |
| Johnny Harris | 42 | 41 | 88% | 93% |
| PBS Space Time | 20 | 27 | 45% | 56% |
| SciShow | 23 | 23 | 49% | 82% |
| **Total** | **109** | **101** | **45%** | **57%** |

## D. Keyword Extraction

The KeyBERT model returned 3103 keywords from the transcripts. After grouping using `gpt-4o-2024-08-06`, this was reduced to 1241 keywords. Finally, after another round of grouping, 9 succinct categories surfaced. Similarly, KeyBERT returned 1020 keywords from the sponsored ad segments detected, `gpt-4o-2024-08-06` reduced this to 377, and the final grouping step further reduced this to 4. The categories and their prevalence are summarized in Table 3.

The content and ad categories are illustrated in Figure 4. Despite being educational channels (as confirmed by the content categories), most of the videos included sponsored ad segments related to products. *Mathematics* videos and those classified as *Various* had almost all sponsored ads related to products. *Physics* videos almost entirely promoted science-related media such as *Nebula* or other streaming services. The other categories included *Education*, *Media*, *Product*, and *Various* sponsored ad segments.

The high prevalence of *Product* sponsored ad segments may indicate that such companies focus heavily on sponsoring YouTube content creators to promote their products compared to other types of companies.

TABLE III. KEYWORDS IN CONTENTS AND SPONSORED AD SEGMENTS

| Content | | Sponsored Ad Segment | |
|---|---|---|---|
| *Category* | *Count* | *Category* | *Count* |
| Architecture | 73 | Education | 58 |
| Biology | 129 | Media | 36 |
| Geopolitics | 33 | Product | 89 |
| Mathematics | 65 | Various | 21 |
| Nature | 27 | | |
| Physics | 54 | | |
| Space | 79 | | |
| Technology | 23 | | |
| Various | 140 | | |

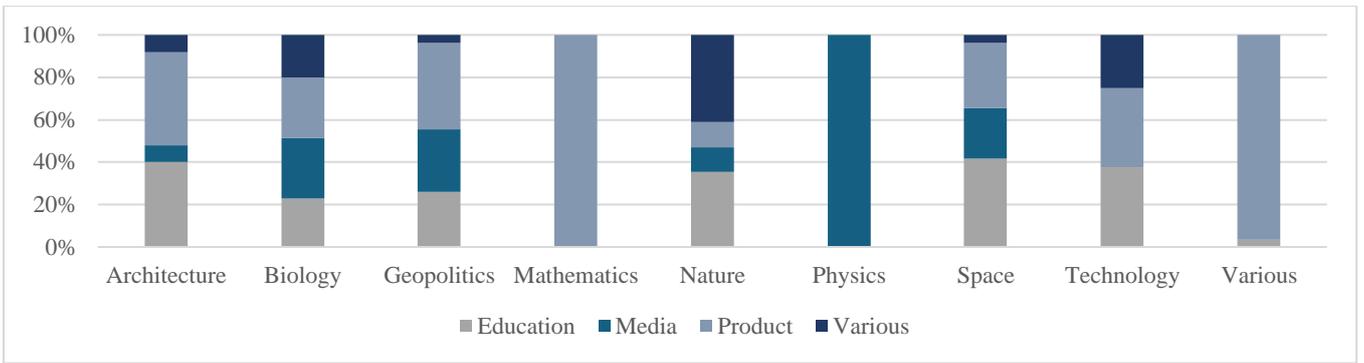

Fig. 4. Content categories compared to sponsored ad segment categories.

## VI. DISCUSSION

This section provides a discussion of the results regarding the prevalence of ads in educational YouTube videos, their relevance to the video content, the distribution of ad types, and the implications of rapid prototyping using ChatGPT.

### A. Prevalence and Relevance of Sponsored Ad Segments

The findings reveal a significant presence of sponsored ad segments across the analyzed videos. This prevalence suggests that even within educational contexts, content creators actively seek partnerships with sponsors to monetize their channels. This trend highlights a growing trend among content creators to integrate advertising as a crucial revenue stream given the direct correlation between sponsored product ad segments and potential viewer purchases [25], [26].

The analysis further underscores the importance of contextual relevance in advertising, particularly within educational content. The ability of sponsors to connect their products with specific topics or subject matter enhances the likelihood of viewer engagement and acceptance. For instance, the significant presence of media-related ads in *Physics* videos, such as promotions for *Nebula* or other science streaming services, illustrates an effective alignment between content and ads. This contextually relevant advertising benefits the advertisers through targeted outreach and enhances the viewer experience by providing related content suggestions.

The findings raise important considerations regarding the quality and integrity of educational content. While some ads may complement the educational material, others could detract from the learning experience if perceived as irrelevant or intrusive. This misalignment can lead to viewer disengagement, negatively affecting creators' and advertisers' experience [10], [11].

### B. ChatGPT and Sponsored Ad Segment Detection

The integration of ChatGPT into the ad detection and keyword extraction process has demonstrated significant effectiveness in identifying sponsored ad segments within video transcripts and generating succinct categories.

Unlike traditional methods, which may struggle to detect subtle forms of advertising, ChatGPT was able to recognize linguistic patterns and contextual cues that indicate the presence of an ad. This capability is important in videos where product placements or sponsorship mentions may be integrated into the narrative without clear demarcation.

ChatGPT flagged sponsored segments within transcripts, allowing a more comprehensive understanding of the advertising landscape across video categories. This approach enhanced the accuracy of ad detection and provided insights into how ads are woven into educational content, contributing to a more nuanced analysis of viewer engagement.

In addition to identifying ads, ChatGPT has proven effective in generating succinct categories from the extracted keywords. The initial extraction process using KeyBERT was then refined through ChatGPT and distilled into 9 final categories for the contents transcripts and 4 categories for the ads transcripts. This ability to condense large data sets into coherent and relevant categories allows for identifying overarching themes that emerge from the keywords, enabling researchers to categorize the data meaningfully.

The integration of ChatGPT into this research has also proven invaluable for rapid prototyping and iterative development. Its use for ad detection and grouping keywords with no fine-tuning allowed for the swift processing of vast amounts of data, significantly reducing the time and effort typically required for manual categorization or developing specialized models.

Furthermore, the ability to scale this approach means that similar methodologies can be applied to other datasets and contexts, expanding the potential applications of this research framework. As the digital landscape continues to evolve, the role of LLMs like ChatGPT in enhancing data analysis capabilities will likely become increasingly significant. By embracing the rapid prototyping capabilities afforded by these models, researchers can navigate the complexities of AI model development much faster [15].

## VII. LIMITATIONS & CONTINUED WORK

While this research offers valuable insights into the relationship between sponsored ad segments and video content, several limitations must be acknowledged:

The dataset consists of approximately 421 video transcripts from educational YouTube channels. While this provides a foundational analysis, the relatively small sample size may limit the generalizability of the findings. Furthermore, the focus on specific channels may not adequately represent the diversity of content available on the platform, which could result in biased insights regarding ad types and relevance. Collecting data on videos with potentially controversial sponsored ad segments would also enable the development of specialized AI models that can reduce harm and augment content restriction rules [27], [28].

The identification of ads and categorization of keywords heavily rely on the model's contextual understanding. While ChatGPT is effective in discerning linguistic cues, the possibility of misinterpretation remains. Certain ads may not be overtly stated, leading to potential misidentification.

How ads are integrated into educational content can vary widely across creators and genres. Some ads may seamlessly incorporate into the narrative, while others might appear as more traditional interruptions. This variability poses challenges in establishing a standardized approach to identifying and evaluating ad relevance, as the impact on viewer engagement may differ based on the integration style.

The dataset labels, including those related to ad detection and keyword categorization, have not been verified by humans. This reliance on automated processes may introduce inaccuracies in identifying and classifying sponsored ad segments, as the model may not accurately capture more nuanced elements. Human verification is crucial for ensuring the labels' validity and the findings' overall reliability.

We expect to continue our research to tackle these limitations. The human verification of the ad detection and categories is crucial to ensuring the validity of the dataset we aim to create. We intend to manually label a sample of the data collected to be used as ground truth for evaluation using metrics such as precision, recall, and F1-scores. This will provide a more rigorous assessment of the system's effectiveness. In addition, our current research has focused on `gpt-4o-2024-08-06` – we plan on continuing our work and implementing a similar pipeline using other OpenAI models such as `gpt-4-turbo-2024-04-09`.

Future research could also explore viewer perceptions of relevance in sponsored segments and its correlation with overall viewer satisfaction and engagement metrics. Another interesting avenue would be whether we can detect subtle messaging or ads indicating a particular channel bias towards a certain product or company, impacting transparency [24]. Furthermore, to make our system more generalizable, we plan to collect transcripts from other video topics, such as entertainment, lifestyle, gaming, food, health, and technology. This will provide a more holistic assessment of our sponsored ad segment detection system.

VIII. CONCLUSION

This research in-progress provides a detailed analysis of the relationship between sponsored ad segments and educational video content on YouTube, based on approximately 421 video transcripts. The findings reveal a significant presence of product-related ads, indicating that advertisers strategically target educational content to reach audiences. By leveraging advanced tools like ChatGPT for ad detection and keyword extraction, the study enhances the analysis of how ads are integrated into YouTube videos.